\documentclass[runningheads]{llncs}
\usepackage[T1]{fontenc}
\usepackage{graphicx}
\usepackage{amsmath}
\usepackage{wrapfig}
\usepackage{booktabs}
\usepackage{appendix}
\usepackage[table]{xcolor}

\definecolor{dustygreen}{rgb}{0.60, 0.74, 0.62}
\definecolor{dustypink}{rgb}{0.87, 0.67, 0.70}
\definecolor{darkerdustygreen}{rgb}{0.45, 0.65, 0.50}
\definecolor{darkerdustypink}{rgb}{0.6, 0.35, 0.38}

\begin{document}
\title{HalCECE: A Framework for Explainable Hallucination Detection through Conceptual Counterfactuals in Image Captioning}

\author{Maria Lymperaiou*\orcidID{0000-0001-9442-4186} \and
Giorgos FIlandrianos*\orcidID{0000-0002-7015-7746} \and
Angeliki Dimitriou*\orcidID{0009-0001-5817-3794} \and Athanasios Voulodimos \orcidID{0000-0002-0632-9769} \and Giorgos Stamou\orcidID{0000-0003-1210-9874}}

\authorrunning{M. Lymperaiou et al.}

\institute{National Technical University of Athens \\
\email{\{marialymp, geofila, angelikidim\}@ails.ece.ntua.gr} \\
\email{   thanosv@mail.ntua.gr
}
\email{gstam@cs.ntua.gr}\\
* These authors contributed equally}
\maketitle              
\begin{abstract}
In the dynamic landscape of artificial intelligence, the exploration of hallucinations within vision-language (VL) models emerges as a critical frontier. This work delves into the intricacies of hallucinatory phenomena exhibited by widely used image captioners, unraveling interesting patterns. Specifically, we step upon previously introduced techniques of conceptual counterfactual explanations to address VL hallucinations. The deterministic and efficient nature of the employed conceptual counterfactuals backbone is able to suggest semantically minimal edits driven by hierarchical knowledge, so that the transition from a hallucinated caption to a non-hallucinated one is performed in a black-box manner. HalCECE, our proposed hallucination detection framework is highly interpretable, by providing semantically meaningful edits apart from standalone numbers, while the hierarchical decomposition of hallucinated concepts leads to a thorough hallucination analysis. Another novelty tied to the current work is the investigation of role hallucinations, being one of the first works to involve interconnections between visual concepts in hallucination detection. Overall, HalCECE recommends an explainable direction to the crucial field of VL hallucination detection, thus fostering trustworthy evaluation of current and future VL systems.

\end{abstract}

\section{Introduction}
In the ever-evolving landscape of artificial intelligence, the appearance of hallucinations has emerged as a significant concern. While neural models showcase remarkable linguistic and/or visual prowess and creativity, their outputs occasionally veer into unpredictable directions, blurring the line between factual accuracy and imaginative fabrication. Hallucinations as a research topic have recently received attention in NLP, with Large Language Models (LLMs) producing unfaithful and inaccurate outputs, despite their sheer size in terms of trainable parameters and training data volume \cite{liu2024surveyhallucinationlargevisionlanguage,zhang2023sirens,tonmoy2024comprehensive,bai2024hallucinationmultimodallargelanguage}.

The nature of hallucination is tied with difficulty in their detection for several reasons, one of them being the variability in hallucination types. \cite{zhang2023sirens} recognize three hallucination categories: Input-Conflicting Hallucinations refer to unfaithful LLM generation in comparison to what the input prompt requested. Context-Conflicting Hallucinations involve inconsistencies within the generated output itself. Finally, Fact-Conflicting Hallucinations violate factuality, providing false information in the output.

Even though LLM hallucinations have captivated significant interest in literature, multimodal settings, such as vision-language (VL) hallucinations, have not been adequately explored yet. Especially during the timely transition towards Large VL Models (LVLMs) \cite{liu2023visual,GPT4TR,chu2024visionllamaunifiedllamabackbone,TheC3}, impressive capabilities in VL understanding and generation are unavoidably accompanied by unfaithful outputs that are even more difficult in detection compared to LLM hallucinations due to intra-modality ambiguity and alignment challenges.

\begin{wrapfigure}{r}{0.48\textwidth}
  \begin{center}
  \vskip -1.1cm
    \includegraphics[width=0.45\textwidth]{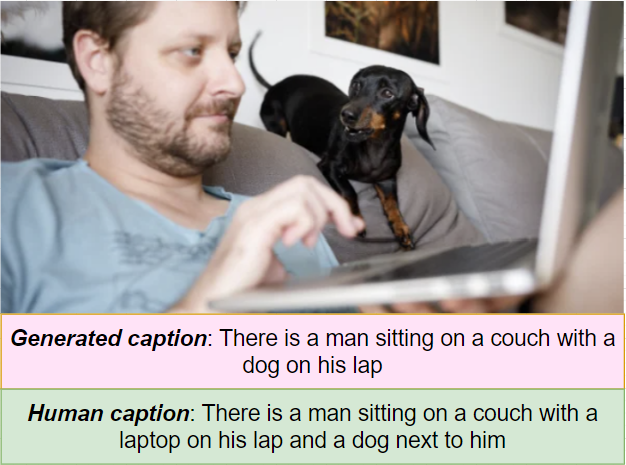}
  \end{center}
  \vskip -0.4cm
\caption{Example of hallucination on image captioning. The generated caption $c$ misses an accurate relationship between the man and the dog. The concept "laptop" should replace the concept "dog" in the generated caption, while the relationship "next to" should be added to connect the concepts "dog" and "man".}
    \label{fig:hal-example}
  \vskip -0.6cm
\end{wrapfigure}

The literature on hallucinations of VL models so far addresses some fundamental research questions regarding evaluation \cite{rohrbach2019object,Li2023EvaluatingOH,wang2023evaluation,jing2023faithscore,lovenia-etal-2024-negative,zhang2024vluncertaintydetectinghallucinationlarge} and mitigation \cite{zhao2023hallucinations,tonmoy2024comprehensive,leng2023mitigating,liu2023mitigating,qu2024alleviatinghallucinationlargevisionlanguage,manevich-tsarfaty-2024-mitigating}. Nevertheless, it grapples with inherent limitations, notably in terms of the interpretability and granularity of metrics employed, hindering a comprehensive understanding of the nuanced challenges posed by hallucinatory phenomena in VL models. We argue that the current VL hallucinations research gaps emphasize the need for an \textit{explainable evaluation} strategy \cite{lymperaiou-etal-2022-towards}, which not only interprets inner workings behind hallucination occurrence, but also paves the way towards effective hallucination mitigation approaches. At the same time, we recognize some related endeavors in recent VL evaluation literature \cite{lymperaiou2023counterfactual}, even though the term "hallucination" is not explicitly used.

In this work, we set the scene for an \textbf{explainable evaluation framework} of VL hallucinations by applying our approach to image captioning, a task associated with hallucination challenges, as in Figure \ref{fig:hal-example}. We borrow techniques from prior work in VL hallucination evaluation, specifically targeting image generation from language \cite{lymperaiou2023counterfactual}, showcasing their effortless applicability in the reverse task of language generation from text. Since most current hallucination evaluation research focuses on object hallucination, i.e. the appearance of extraneous objects or, on the contrary, objects consistently missing in the generated output, while only a few  assess role hallucinations \cite{role-hal}, referring to spatial relationships or actions, we aim to construct a unified framework named HalCECE, incorporating both of them through their projection on \textbf{graph edits}.
In our proposed HalCECE framework, we retain fundamental properties of conceptual counterfactuals \cite{filandrianos2022conceptual} and graph-driven edits \cite{structure-your-data}, which will be analyzed in subsequent sections.
Overall, we present the following contributions:
\begin{itemize}
\item We propose the adoption of explainable evaluation in image captioning hallucination detection contrasting typical captioning evaluation metrics.
\item We decompose concepts existing in captions to allow fine-grained evaluation and quantification of hallucination.
\item We substantiate our findings by applying our proposed evaluation framework to various image captioners of increasing model size.
\end{itemize}

\section{Background}
\paragraph{\textbf{Image captioning}} stands as a pioneering task in machine learning, bridging the visual and linguistic modalities so that an accurate intra-modality communication can be established. 
Real-world AI systems rely on image captioning to provide textual descriptions for visually-impaired people, facilitate indexing and retrieval of images based on textual requests and enable image-language alignment for advanced interaction between humans and computers. To this end, hallucinations arise as a crucial concern that impede the effective usage of visual descriptions in practice \cite{Ghandi2022DeepLA}.

With the advent of VL transformers, the field of image captioning has made substantial strides, with models such as BLIP \cite{li2022blip}, BLIP-2 \cite{Li2023BLIP2BL}, Llava \cite{liu2023visual,liu2023improvedllava}, BEiT \cite{Wang2022ImageAA}, GiT \cite{Wang2022GITAG} and others achieving state-of-the-art results in the low-billion parameter regime. Despite scaling up towards billion trainable parameters \cite{GPT4TR,TheC3,chu2024visionllamaunifiedllamabackbone}, which serves as a general criterion for language generation quality, many image captioners often come across hallucinations in the generated text, the detection and mitigation of which becomes even more challenging under closed-source scenarios, as in the case of GPT-4 \cite{GPT4TR} and Claude \cite{TheC3} models.

\paragraph{\textbf{Hallucinations in VL models}} refer to the appearance of non-existent concepts in the generated modality. 
Evaluation of complex VL systems for language generation from images has largely exploited metrics focusing on linguistic quality, such as BLEU \cite{papineni-etal-2002-bleu}, ROUGE \cite{lin-2004-rouge}, CIDEr \cite{vedantam2015cider} and others, which have been widely used for benchmarking. Association between intra-modality concepts, i.e. the agreement of visual and linguistic cues has only recently been addressed, under the spotlight of VL hallucinations.

Recent research has showcased some interesting endeavors towards capturing hallucinations, targeting different aspects of the problem, and employing varying techniques.
With a focus on objects, detection of VL hallucinations was initially performed using the CHAIR (Caption Hallucination Assessment
with Image Relevance) metric \cite{rohrbach2019object}. The per instance $\textit{CHAIR}_i$ is defined as:
\begin{equation}
\textit{CHAIR}_i=\frac{|\textit{Hallucinated Objects}|}{|\textit{All Predicted Objects}|}
\end{equation}
Furthermore, the per sentence $\textit{CHAIR}_s$ is formed as:
\begin{equation}
\textit{CHAIR}_s=\frac{|\textit{Sentences with hallucinated objects}|}{|\textit{All sentences}|}
\end{equation}
Despite its simplicity, CHAIR acts as a first, immediate measure for object hallucinations, inspiring more refined consequent approaches.

The FAITHSCORE metric addresses different types of VL hallucinations in a fine-grained way by breaking down the caption in subcaptions, from which atomic facts are extracted \cite{Jing2023FAITHSCOREEH}. Similarly, ALOHa \cite{petryk-etal-2024-aloha} leverages an LLM to identify groundable objects within a candidate caption, assess their semantic similarity to reference objects from both captions and object detections, and utilize Hungarian matching to compute the final hallucination score. Nonetheless, the subcaption process leverages LLMs, which also hallucinate themselves. 

The dialog-based evaluation process of POPE \cite{Li2023EvaluatingOH} suggests answering "yes/no" to questions regarding the existence of an object in an image. Objects are extracted from images based on ground truth annotations or segmentation tools, filling  question templates, while an equally sized set of non-existent objects provides negative samples to measure the confidence of prompted models against "yes/no" answer bias. Then, the agreement between  answers with ground truth objects is measured. 
Also using a question-answering pipeline to evaluate object hallucinations, NOPE \cite{lovenia-etal-2024-negative} regards LLM-constructed questions with negative indefinite pronouns (e.g. nowhere, none etc) as ground truth answers.

Involving LLMs in the hallucination detection pipeline,
\cite{wang2023evaluation} are the first to recognize VL hallucination patterns, driving the construction of prompts for ChatGPT to generate relevant hallucinated instances. Fine-tuning LLama \cite{touvron2023llama} on such hallucinations provides a proficient module for capturing VL hallucinations.

Model performance on standard text generation metrics may be negatively correlated with hallucination occurrence, while the choice of image encoding techniques and training objectives employed in the pre-training stage can be definitive \cite{Dai2022PlausibleMN}.
Statistical factors accompanying object hallucinations were analyzed in \cite{Zhou2023AnalyzingAM}, examining frequent object co-occurrences, uncertainty during the generation process, and correlations between hallucinations and object positioning within the generated text.

\section{Explainable hallucinations evaluation}
\label{sec:hal-eval}
Many of the contributions analyzed above harness LLMs at some point of the hallucination evaluation process. These approaches inevitably induce uncertainty related to the prompt used, while simultaneously facing the possibility of LLMs also hallucinating and ultimately hindering the robustness and trustworthiness of the affected module, and thus the evaluation framework itself. In our framework, we deviate from the usage of LLMs, sacrificing the simplicity they provide in order to enhance the determinism and reliability of the evaluation process.

Other than that, both metrics evaluating linguistic quality, as well as metrics for VL hallucinations lack explainability aspects, since they do not suggest the \textit{direction of change} towards dehallucinated generation. This direction of change should primarily be \textbf{measurable} and \textbf{meaningful}, while its optimal usage prescribes notions of \textbf{optimality}, translated to semantically \textbf{\textit{minimal changes}}, as well as the \textbf{\textit{fewest possible number of edits}} leading to the desired outcome. We will analyze these desiderata:

\paragraph{Measurable} change refers to assigning a well defined numerical value for comparative reasons. This requirement demands the connection of concepts to be changed with similarity features within a unified structure, such as their distance on a semantic space or within a semantic graph.
\paragraph{Meaningful} change refers to performing operations that are sensible in the real world, such as substituting an object with another object and not with meaningless sequences of characters. For example, swapping the concept "cat" with the character sequence "hfushbfb" does not hold a useful meaning. Moreover, even substituting objects with actions breaks meaningfulness, e.g. replacing the concept "cat" with the concept "swimming" within the same sentence violates the well-defined rules of linguistic syntax.
\paragraph{Optimal} change refers to employing a strategy which guarantees that valid and measurable changes are the best ones to be found among a possibly infinite set of valid and measurable changes. For example, replacing "cat" with "person" is meaningful for a human, while also being measurable if we place the concepts "cat" and "person" in a semantic graph structure. However, an alternative suggestion could be replacing "cat" with "dog", as they are both animals, or even "cat" with "tiger" since they are both felines. In this case, optimal edits require finding the most \textit{semantically similar concept} to the source one. Such similarity requirements can be imposed by structured knowledge bases, deterministically ensuring \textbf{\textit{semantically minimal edits}}. Furthermore, the number of such edits should be controlled, since infinitely performing minimal changes should be naturally excluded from the proposed framework. For example, the transition "cat"$\rightarrow$"dog" should not consider extraneous changes, if not required to approach the ground truth sample. Therefore, the set of all proposed changes should be minimized in terms of overall semantic cost, ultimately resulting in \textbf{\textit{fewest possible semantically minimal edits}}.

To address these challenges, we leverage the framework first proposed in \cite{filandrianos2022conceptual}, where counterfactual explanations are provided via edits satisfying our desiderata. This framework was later adopted for the evaluation of image generation models \cite{lymperaiou2023counterfactual}, where a source set $S$ contains the ground truth concepts as extracted from the generated modality (in our use case being textual captions) and a target set $T$ contains ground truth concepts as extracted from the input modality (in our case being annotated images provided to the captioner).

We wish to perform the $S\rightarrow T$ transition using the fewest possible semantically minimal and meaningful edits, which is achieved via the guarantees offered by the WordNet hierarchy \cite{miller1995wordnet}: concepts from $S$, $T$ are mapped on WordNet synsets, which correspond to sets of cognitive synonyms. Distances of synsets within the hierarchy translate to semantic differences in actual meaning. Finding the minimum path between two synsets entails semantically minimal differences between corresponding concepts. WordNet is a crucial component of this implementation, since it guarantees \textbf{measurable} (WordNet distance is a numerical value), \textbf{meaningful} (WordNet synsets correspond to lexical entities of the English language) and \textbf{semantically minimal} (shortest WordNet distance between two concepts is found using pathfinding algorithms \cite{dijkstra1959note}) concept edits.
The algorithm of \cite{filandrianos2022conceptual} uses bipartite matching to minimize the overall cost of assignment between $S$ and $T$ concepts, ensuring the optimal $S\rightarrow T$ transition.

By breaking down the $S\rightarrow T$ transition, the following three edit operations $e$ are allowed for any source  $s\in S$ and target concept $t\in T$ \cite{filandrianos2022conceptual,lymperaiou2023counterfactual}:
\begin{itemize}
    \item \textbf{Replacement (R)} $e_{s \rightarrow t}(S)$: A concept $s \in S$ is replaced with $t  \notin S$.
    \item \textbf{Deletion (D)} $e_{s-}(S)$: A concept $s \in S$ is deleted from $S$.
    \item \textbf{Insertion (I)} $e_{t+}(S)$: A concept $t \in T$ is inserted in $S$.
\end{itemize}

Especially in the case of image captioning, we impose higher importance in \textbf{D} and \textbf{R} edits; the rationale behind this decision is that since hallucinations refer to the presence of irrelevant or extraneous concepts, they should be deleted or replaced to match the ground truth ones. Moreover, in many cases, captions purposefully provide a higher-level description of an image, therefore several visual concepts are omitted, sacrificing coverage for brevity. In that case, \textbf{I} suggests the addition of visual concepts to the caption, which may not be always necessary. In our framework, we also include \textbf{I} calculations, but we do not consider them in the overall transformation cost; instead, we provide them as \textit{suggestion} for the user to choose whether they may be incorporated in more verbose captions.  
\subsection{The role of roles}
In Figure \ref{fig:hal-example}, the captioning model (BLIP) confuses the spatial relationship between the man and the dog, showcasing the importance of \textit{role hallucinations}, which were not widely addressed in prior work, since object hallucinations were their primary concern. Additionally, roles should be addressed\textit{ in conjunction to objects}, and not on their own, since this more simplistic approach would result in under-detection of hallucinations. For example, if we apply the counterfactual explanations algorithm of \cite{filandrianos2022conceptual} on sets of roles, the proposed edits for Figure \ref{fig:hal-example} would be \{\textbf{I}("next to")\}, referring to the addition of the role "next to" that connects the dog and the man. However, if we consider \textit{triples} of two objects connected with a role, the resulting edits would be: \{\textbf{R}(["dog", "on", "lap], ["laptop", "on", "lap"]), \textbf{I}(["dog", "next to", "man"])\}, which is a more valid set of edits, if we view the human-written ground truth caption and the image itself.

To perform the transition to editing triples instead of standalone concepts, we require scene graphs instead of objects to acquire a conceptual representation of the image \cite{structure-your-data}. Regarding the caption, we also parse the sentence in a graph structure.
Given two graphs $G_T$ representing the image and $G_S$  corresponding to a possibly hallucinated generated caption $c$, we search for the minimum cost set of \textbf{R}, \textbf{D}, \textbf{I} edits (applied on objects and roles) that transform $G_S\rightarrow G_T$, i.e. convert a -possibly- hallucinated graph to a non-hallucinated one.

This cost of transformation is calculated using Graph Edit Distance (GED) between $G_S$, $G_T$. Denoting as $c(e_i)$ the cost of an operation $e_i \in \{\textbf{R}, \textbf{D}, \textbf{I}\}$ and $P(G_S, G_T)$ the set of $n$ edit paths to transform $G_S\rightarrow G_T$, GED is formed as:
\begin{equation}
    \textit{GED($G_S$, $G_T$)}=\min_{(e_1,..., e_n)\in P(G_S, G_T)}\sum_{i=1}^{n}c(e_i)
\label{eq:eq1}
\end{equation}
The shortest paths $P(G_S, G_T)$ are calculated using deterministic pathfinding algorithms, such as Dijkstra \cite{dijkstra1959note}, ensuring optimality of edits $e_i$.

However, GED is an NP-hard algorithm, meaning that it cannot be calculated efficiently in its basic, brute-force format. For this reason, we employ some approximations, such as the Volgenant-Jonker (VJ) algorithm \cite{jonker1987shortest}, which allows for GED calculation in polynomial time.

\begin{figure}[t]
\centering\vskip -0.1cm
\includegraphics[width=1.03\textwidth]{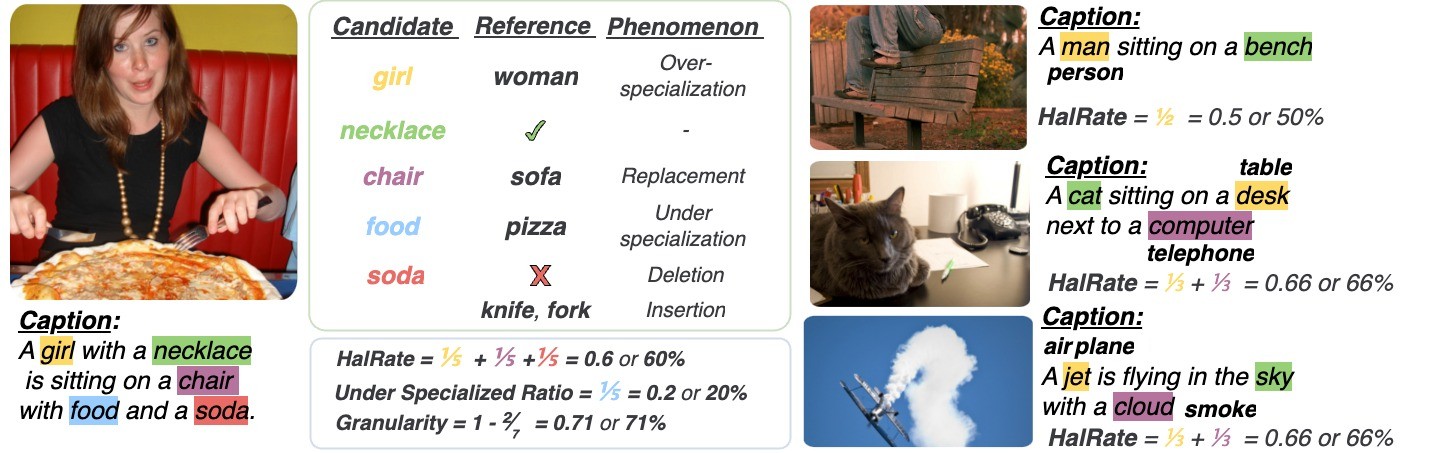}
\caption{An example of detected hallucination of objects in image captioning from our framework is presented, depicting each phenomenon along with the proposed metrics. Objects in yellow represent an overspecialized phenomenon, in purple a replacement, and in red a removal. Those in green are correct objects, and those in blue are the underspecialized objects (which do not constitute hallucinations, as the caption contains a more generic concept to the ground truth one). As shown, the hallucination rate is calculated as the sum of the rate of each hallucination phenomenon independently.} 
\label{fig:demo-hal}
\vskip -0.1cm
\end{figure}

\section{Hallucination detection framework}
\label{sec:hal-detect}

\paragraph{\textbf{Object hallucinations}}
In Figure \ref{fig:demo-hal} we illustrate the hallucination analysis provided by our framework. We formulate the object hallucination detection problem as follows: each caption $c$ has generated objects $S = \{s_1, s_2, \ldots, s_n\}$, and each image contains the ground truth objects $T = \{t_1, t_2, \ldots, t_m\}$. We find the \textbf{R}, \textbf{D}, \textbf{I} sets of object edits to perform the transition $S\rightarrow T$, as analyzed in §\ref{sec:hal-eval}. 

In order to evaluate different granularities of hallucinations, i.e. presence of more generic or more specific concepts compared to the ground truth one, we utilize the Least Common Ancestor (LCA) within the WordNet hierarchy. Specifically, LCA denotes the closest ancestor synset between two synsets in WordNet; we closely examine the case where the LCA between two synsets contains one of the synsets itself: for example, given two synsets $v$ and $w$, if LCA($v$, $w$)=$v$, then $v$ is a hypernym (more generic concept) of $w$.

Based on these, we analyze the following \textit{hallucination phenomena}:
\begin{itemize}
    \item \textbf{Deletion (D)}: When an object $s_i\in S$ must be deleted; e.g., in Figure \ref{fig:demo-hal}, the concept "soda" is in the generated caption $c$ but not in the image.
    \item \textbf{Replacement (R)}: When an object $s_i \in S$, is replaced with a different object $t_j$, where $LCA(s_i, t_j) \neq s_i$, and $LCA(s_i, t_j) \neq t_j$ (meaning that no object is a hypernym of the other). For instance, the caption references a "chair", but the image contains a "sofa".
    \item \textbf{Over-specialization (O)}: When an object $s_i \in S$ is replaced with a different object $t_j$, where $LCA(s_i, t_j) = t_j$, i.e. $t_j$ is a more general concept than $s_i$ in the hierarchy. For example, the caption states that the image contains a "girl", but the image depicts a "woman"; in this case, the caption erroneously overspecified this term, since "girl" is subcategory of "woman".   
\end{itemize}

Based on these phenomena, we measure the degree of hallucination for a caption $c$ as the number of objects that exhibit at least one of the aforementioned phenomena. Thus, the metric for counting hallucinations in captioning, denoted as $\textit{Hal}(S, T)$, is defined as the sum of the cardinalities of the sets of \textbf{D}, \textbf{R}, \textbf{O}:
\begin{equation}
    \textit{Hal}(S, T) = \lvert \textbf{D}(S, T) \rvert 
    + \lvert \textbf{R}(S, T) \rvert \\
    + \lvert \textbf{O}(S, T) \rvert
\end{equation}
The hallucination rate $\textit{HalRate}$ reveals the percentage of hallucinated objects over the total number of objects $\lvert S \rvert$ in $c$, and it is mathematically expressed as:
\begin{equation}
\begin{aligned}
    \textit{HalRate}(S, T) = \frac{\textit{Hal}(S, T)}{\lvert S \rvert} 
\end{aligned}
\end{equation}
We incorporate additional semantic metrics on these properties, such as quantifying the semantic distance between hallucinatory and ground truth concepts. 
\begin{itemize}
    \item \textbf{Similarity of Replacements}: We employ Wu-Palmer similarity \cite{wu1994verb} to measure the semantic similarity of replacements based on the position of synsets in WordNet. This way, we measure how close the replaced terms are in order to gain further understanding of the behavior of the captioner. For example, semantically related replacements receive a higher Wu-Palmer similarity score, denoting more "justified" hallucination occurrences.
\end{itemize}
An additional facet of HalCECE lies in its capacity to explore phenomena beyond hallucination. This is exemplified through the following measures:
\begin{itemize}
    \item \textbf{Granularity}:
    Defined as $1$ minus the ratio of \textbf{Insertions (I)} over the number of ground truth image objects.     In essence, it represents the percentage of objects that $c$ attempts to encapsulate compared to the image objects:
\begin{equation}
            Granularity(S, T) = 1 - \frac{\lvert \textbf{I}(S) \rvert}{\lvert T \rvert} 
    \end{equation}
\item \textbf{Under-Specialization (U)}: Quantifies the instances of underspecialized objects, where an object $s_i \in S$ from $c$ is replaced with a different object $t_j$, and 
$LCA(s_i, t_j) = s_i$, meaning that the caption object is more generic than the corresponding image object. For instance, if the caption indicates the presence of "food", but the image portrays a "pizza", the caption is not incorrect (because a "pizza" is a sub category of "food") but could benefit from greater specificity. The ratio is computed as the division of the number of under-specialized objects by the total number of objects in  $c$, reflecting the proportion of objects in the generated captions that are underspecialized.
\end{itemize}
In our analysis, we incorporate both the \textbf{average number of objects per caption} and the \textbf{average number of WordNet ancestors (hypernyms)} associated with each of these objects for all data instances. This approach provides a comprehensive perspective on the content of each caption $c$.

\paragraph{\textbf{Role hallucinations}}
Our framework is directly extended to incorporate edge-level hallucinations. On top of objects included in $T$ and $S$, images and captions also describe object interactions. As explained, role hallucination is measured using \textit{triples} and not simply relations which would disregard adjacent objects and their transformations. To this end, we denote the sets of triples corresponding to captions and image annotations respectively as $S^r = \{(s_i, r^s_j, s_k), \ldots \}$ and $T^r = \{(t_i, r^t_j, t_k), \ldots \}$. A visual representation of roles within captions can be found in Figure \ref{fig:example-roles}. Examples of caption triples are "horse \textit{over} obstacle" and "people \textit{sitting at} table".  To measure role hallucinations, i.e. the transition from $S^r \rightarrow T^r$, we employ an adjusted version of aforementioned equations. Edit sets \textbf{D} and \textbf{R} are calculated by considering triples instead of objects as following:
\begin{figure}[h!]
\centering   \vskip -0.1cm
 \includegraphics[width=0.65\textwidth]{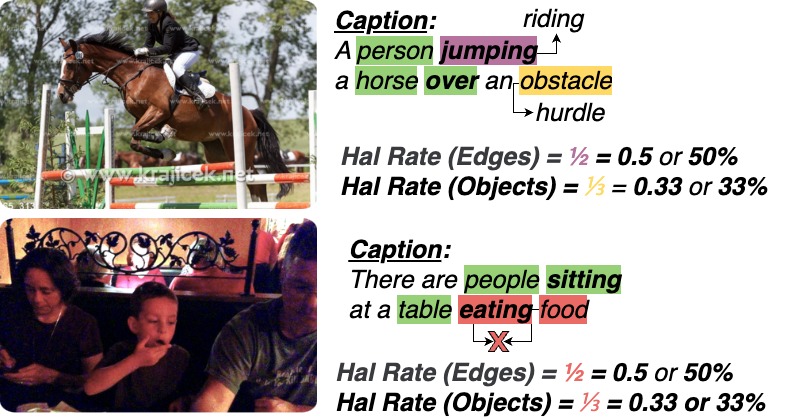}
\caption{An demonstration of the edge integration into HalCECE. The edges are highlighted in \textbf{bold}, and the different colors correspond to those of Figure \ref{fig:demo-hal}.}
\label{fig:example-roles}
  \vskip -0.5cm
\end{figure}
\begin{itemize}
    \item \textbf{Deletions (D)}: When an edge $r^s_j$ between two objects $s_i$, $s_k$ must be deleted. Notably, this edit set includes deletions induced as "collateral damage" due to object deletions or replacements, as well as hallucinated relations between correctly detected objects in $c$. In Figure \ref{fig:example-roles}, the role "eating" between "people" and "food" is deleted  because "food" is hallucinated by the captioner.
    \item \textbf{Replacement (R)}: When an edge $r^s_j$ between two objects $s_i$, $s_k$ is replaced with another edge $r^t_w$. For example, in Figure \ref{fig:example-roles} the role "jumping" between "person" and "horse" is hallucinated and needs to be replaced with "riding". Despite "jumping" being a valid relation between "horse" and "obstacle", or even "person" and "obstacle", it is definitely not correct in the presented configuration, placing a great focus on leveraging roles as part of a triple.
\end{itemize}
It is noteworthy that the definition of over-/under- specialization is not applicable for roles, as edges describe actions, topology or "part of" relations, steering away from hierarchies. To combat this, we leverage the annotation information provided by humans to correctly match caption relations to ground truth ones and map them to WordNet. When captioners produce previously unseen relations (in terms of ground truth), we weight them accordingly, so that they can be easily inserted or deleted during GED computation; they are not likely to be replaced with other roles though, since we lack semantic content. To detect if they are part of \textbf{R}, we deploy an extra post-hoc reasoning step and check if  a relation $r^s_j$ between the same two objects has been deleted and another $r^t_w$ has been added.
Given the previous analysis, role hallucinations are measured as:
\begin{equation}
\begin{aligned}
    \textit{Hal}(S^r, T^r) = \lvert \textbf{D}(S^r, T^r) \rvert 
    + \lvert \textbf{R}(S^r, T^r) \rvert \\
\end{aligned}
\end{equation}
while $\textit{HalRate}$ and $\textit{Granularity}$ are simply adjusted to be:
\begin{equation}
\begin{aligned}
    \textit{HalRate}(S^r, T^r) = \frac{\textit{Hal}(S^r, T^r)}{\lvert S^r \rvert} 
\end{aligned}
\end{equation}
\begin{equation}
        \begin{aligned}
            \textit{Granularity}(S^r, T^r) = 1 - \frac{\lvert \textbf{I}(S^r) \rvert}{\lvert T^r \rvert} 
        \end{aligned}
    \end{equation}
\section{Experiments}
\paragraph{\textbf{Dataset and models}} To evaluate HalCECE on images connected with both captions and scene graphs, we experiment on the intersection of Visual Genome (VG) \cite{Krishna2016VisualGC} and COCO \cite{lin2015microsoft}. VG contains handcrafted scene graph annotations incorporating objects, attributes and roles. On the other hand, COCO scenes are connected with 5 captions per image, provided by humans. We restrict our experimentation on the COCO validation set (splits are provided by the dataset creators), which demonstrates 2170 common instances with VG; a few of those are eliminated, if the corresponding objects cannot be aligned with WordNet.

We initially experiment with non-proprietary captioners, evaluating both smaller and larger models, since smaller ones can be more easily deployed by every researcher. Specifically, we apply our method on  variants of GiT \cite{Wang2022GITAG}  and BLIP \cite{li2022blip,blip2}, namely
\textit{GiT-base }(trained on 10 million image-text pairs), \textit{GiT-large} (trained on 20 million image-text pairs) and \textit{GiT-base/large-coco} (fine-tuned on COCO captions); also
\textit{BLIP-base} (using ViT \cite{vit} base encoder),  \textit{BLIP-large} (ViT large encoder), \textit{BLIP2-flan-t5-xl} (Flan-T5 \cite{chung2022scalinginstructionfinetunedlanguagemodels} is used as the language decoder) and \textit{BLIP2-opt-2} (using OPT \cite{zhang2022optopenpretrainedtransformer} 2.7B as the language decoder). We attempt unconditional and conditional image captioning (related experiments will be denoted as \textit{unc/cond}), where captioners are fine-tuned to estimate conditional and unconditional distributions over captions respectively \cite{kornblith2023guiding}. Moreover, we experiment with \textit{ViT-GPT2} \cite{vit-gpt2}, which leverages ViT as the encoder and GPT2 \cite{gpt2} as the decoder.
Finally, we provide results on two proprietary foundational models of the Claude family \cite{TheC3} prompted for captioning, namely \textit{Claude-sonnet}\footnote{anthropic.claude-3-5-sonnet-20241022-v2:0
} and \textit{Claude-haiku}\footnote{anthropic.claude-3-haiku-20240307-v1:0}. This way, we prove the real power of HalCECE on closed-source models where our white-box competitors are not applicable. All parameter counts for these models are detailed in Appendix \ref{sec:parameters}.

Since prompting LVLMs can define the length of the generated captions, we attempt to generate both longer captions (20-30 words), as well as shorter ones (10 words max), which are comparable to the captions produced from the rest of the captioners. This way, we get the opportunity to explore HalCECE on longer descriptions, something that is not available in smaller VL models. We name the respective experiments using \textit{L} for \textit{long} generations and \textit{S} for \textit{short} ones.

\paragraph{\textbf{Concept sets construction}}
We construct the linguistic $S$, $S^r$ and  visual concept sets $T$, $T^r$, corresponding to source and target concept sets respectively with the goal of transforming $S \rightarrow T$ and $S^r \rightarrow T^r$. Linguistic sets are formed by extracting graphs from text via the Scene Graph Parser tool\footnote{https://github.com/vacancy/SceneGraphParser}, while visual sets are constructed using ground truth annotations from COCO and VG. 
\paragraph{\textbf{Experimental setup}} Non-proprietary pre-trained captioners are loaded from Huggingface\footnote{https://huggingface.co/models?pipeline\_tag=image-to-text}  using their respective model cards and their inference is executed on a 12GB NVIDIA TITAN Xp GPU. No further training is performed. Proprietary Claude models are accessed  via Amazon Web Services (AWS) using API calls (Bedrock service). Prompts for Claude models are presented in App. \ref{sec:prompts}.
\subsection{HalCECE Results}
\begin{figure}[t!]
    \centering
\includegraphics[width=\linewidth]{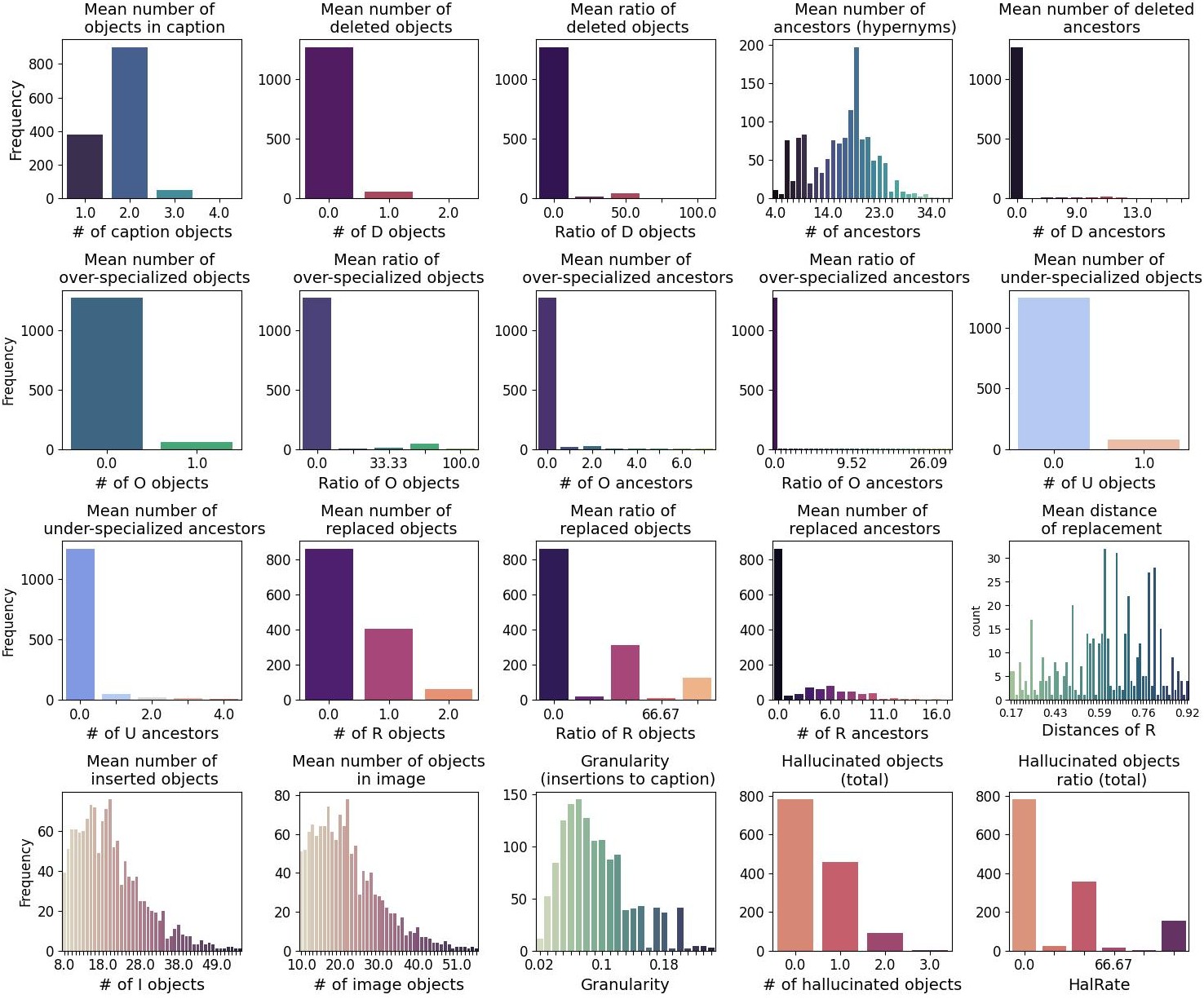}
    \caption{Statistics of our proposed explainable metrics on \textbf{\textit{object hallucinations}} by GiT-base on the $VG \cap \textit{COCO}$ validation set.}
    \label{fig:obj-results}
\end{figure}

\begin{figure}[h!]
    \centering
\includegraphics[width=\linewidth]{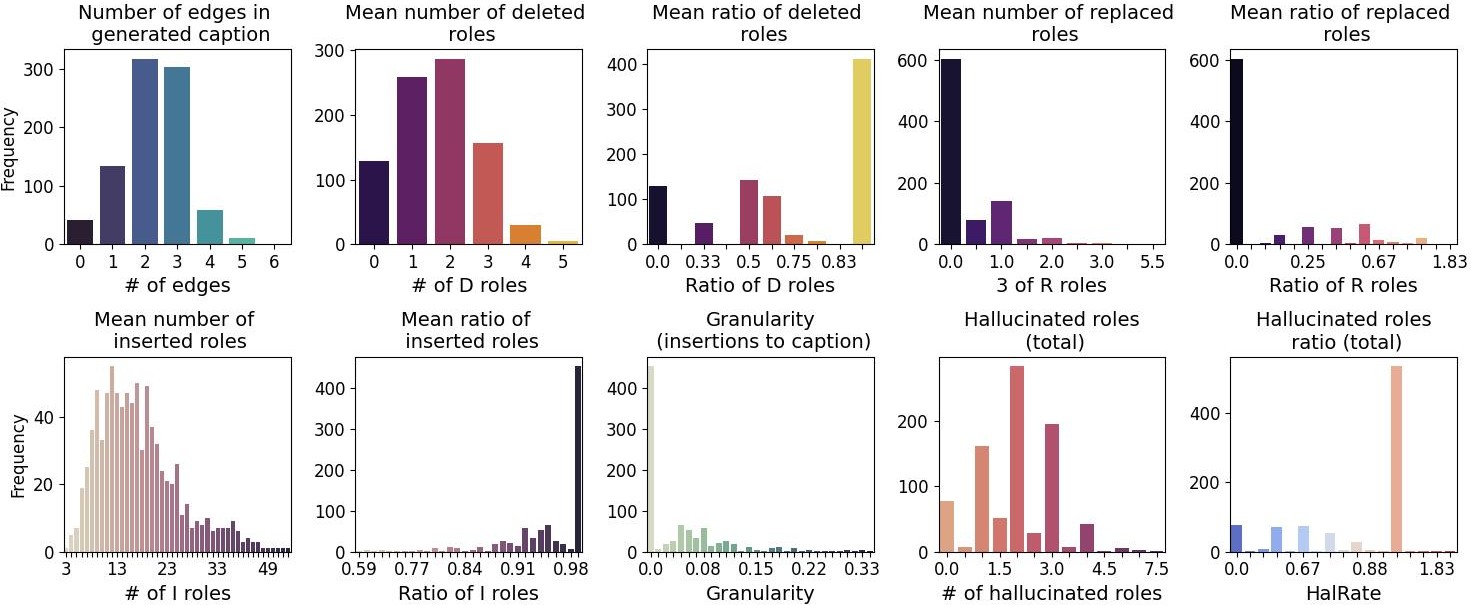}
    \caption{Statistics of our proposed explainable metrics on \textbf{\textit{role hallucinations}} by GiT-base on the $\textit{VG}\cap \textit{COCO}$ validation set.}
    \label{fig:role-results}
\end{figure}

Based on the hallucination detection framework analyzed in the previous section, we present our findings as following: Tables \ref{tab:obj-hallucinate}, \ref{tab:obj-hallucination-2}, \ref{tab:exp-role} contain averaged results per captioner involving the hallucination phenomena introduced above. In addition, Figures \ref{fig:obj-results}, \ref{fig:role-results} demonstrate the distributions of values per hallucination phenomenon in our dataset, addressing object and role hallucinations respectively. These plots refer to GiT-base as a proof-of-concept, since it is one of the best-performing captioners according to our reported explainable metrics.

\begin{table}
\caption{Object hallucinations (mean values) on the $\textit{VG}\cap \textit{COCO}$ validation subset. \textcolor{darkerdustygreen}{\textbf{Best}} and \textcolor{darkerdustypink}{\textbf{worst}} results are denoted. Numbers in parenthesis denote absolute \#objects.}
\label{tab:obj-hallucinate}
\centering \small
\begin{tabular}{p{2.4cm}|>{\centering\arraybackslash}p{1.3cm}|>{\centering\arraybackslash}p{1.6cm}|c|c|>{\centering\arraybackslash}p{1.8cm}}
\hline
\textbf{Model}                                       & \#objects  & \#ancestors   & HalRate (\#hal. objects)$\downarrow$  & Granul. & \textbf{U}$\downarrow$ \\ \hline 
  GiT-base-coco            & 3.13               & 27.93                 & 35.56\% (1.13)          & 17.0\%      & 4.06\% (0.13)     \\
  GiT-large-coco           & 3.15               & 27.97                 & 33.93\% (1.1)           & 17.0\%      & 3.92\% (0.12)     \\
  GiT-base & 1.76               & 16.57                 & 26.41\% (0.48)          & 9.0\%     & 3.27\% (0.06)    \\
  GiT-large                & 1.74               & 16.28                 & \cellcolor{dustygreen}25.38\% (0.46)          & 9.0\%     & 3.31\% (0.06)     \\\hline
 BLIP-base-unc& 2.53               & 22.55                 & 34.28\% (0.91)          & 13.0\%      & 4.48\% (0.12)     \\
 BLIP-base-cond	& 3.23 &	29.5 &	58.48\% (1.87)	& 17.0\% &	2.96\% (0.1) \\

 BLIP-large-unc & 3.63               & 32.73              & 39.2\% (1.45)          & 19.0\%    & 3.47\% (0.13) \\
 BLIP-large-cond	& 4.22 &	37.5 &	53.04\% (2.24) &	22.0\%	& \cellcolor{dustygreen}2.84\% (0.12) \\

BLIP2-flan-t5-xl &	2.57 &	23.16	& 33.13\% (0.89)	& 14\%	& 4.05\% (0.11) \\

BLIP2-opt-2	& 2.78 &	24.89	& 33.28\% (0.96)	& 15.0\% &	4.19\% (0.12) \\
\hline
ViT-GPT2 &	2.95 &	26.51 &	38.76\% (1.18)	& 16.0\%	& 4.47\% (0.14)
\\\hline
Claude sonnet-L & 6.85 &	58.94	& 58.91\% (4.05) &	36.0\%	& 4.71\% (0.33) \\
Claude haiku-L & 7.12	& 58.66 &	\cellcolor{dustypink}64.31\% (4.64) &	39.0\%	& 5.4\% (0.39) \\
Claude sonnet-S & 3.35 &	30.48	& 47.16\% (1.6)	& 17.0\%	& 4.67\% (0.16)\\
Claude haiku-S & 2.95	& 25.49 &	54.36\% (1.62) &	16.0\%	& \cellcolor{dustypink}6.74\% (0.19)\\
\hline
\end{tabular}
\end{table}
\begin{table}[h!]
\caption{Continuation of Tab. \ref{tab:obj-hallucinate}. More object hallucination phenomena on $\textit{VG}\cap \textit{COCO}$ validation subset. Numbers in parenthesis denote absolute \#objects.}
\label{tab:obj-hallucination-2}
\centering \small
\begin{tabular}{p{2.4cm}|c|c|c|>{\centering\arraybackslash}p{2.5cm}}
\hline
\textbf{Model}                                       & \textbf{D}$\downarrow$   & \textbf{O}$\downarrow$  & \textbf{R}$\downarrow$  & Similarity of \textbf{R}$\uparrow$ \\ 
\hline
  GiT-base-coco            & 4.38\% (0.15) & 3.01\% (0.09)    & 28.18\% (0.89) & 0.56                        \\
  GiT-large-coco           & 4.4\% (0.16)  & 2.46\% (0.08)    & 27.06\% (0.87) & 0.55                   \\
  GiT-base   & \cellcolor{dustygreen}2.11\% (0.05) & \cellcolor{dustygreen}2.17\% (0.04)    & 22.12\% (0.4)  & \cellcolor{dustygreen}0.61                        \\
  GiT-large                & 2.46\% (0.05) & 2.41\% (0.04)    & \cellcolor{dustygreen}20.51\% (0.36) & 0.6                         \\\hline
 BLIP-base-unc  & 3.78\% (0.11) & 2.65\% (0.07)    & 27.86\% (0.73) & 0.57  \\
  BLIP-base-cond	& \cellcolor{dustypink}23.07\% (0.72) &	2.76\% (0.09) &	32.66\% (1.05) &	0.52 \\

 BLIP-large-unc & 6.13\% (0.24) & 3.48\% (0.13)    & 29.59\% (1.08) & 0.56     \\
 BLIP-large-cond	& 19.27\% (0.81) &	2.46\% (0.11) &	31.3\% (1.32)	& 0.52 \\

 BLIP2-flan-t5-xl	& 4.27\% (0.12)	& 3.16\% (0.08)	& 25.7\% (0.69) &	0.56 \\
BLIP2-opt-2	& 3.64\% (0.11)	& 2.8\% (0.08)	& 26.84\% (0.77) &	0.57 \\
\hline
 ViT-GPT2	& 3.45\% (0.11) &	3.16\% (0.09) &	32.14\% (0.97) & 	0.6 \\\hline
 Claude sonnet-L & 15.79\% (1.05)	& 2.51\% (0.19) &	40.61\% (2.81) &	0.52\\
Claude haiku-L & 17.3\% (1.28)	& 2.69\% (0.2)	& \cellcolor{dustypink}44.33\% (3.15)  & 	\cellcolor{dustypink}0.49\\
 Claude sonnet-S & 7.1\% (0.25) &	\cellcolor{dustypink}5.42\% (0.18) &	34.63\% (1.16) &	0.57\\
Claude haiku-S & 7.78\% (0.24) &	4.59\% (0.13) &	41.99\%  (1.26)	& 0.52\\
\hline
\end{tabular}
\end{table}
\begin{table}[h!]
\caption{Role hallucinations (mean values per image) on the $\textit{VG}\cap \textit{COCO}$ validation subset. Numbers in parenthesis denote absolute \#roles.}
\label{tab:exp-role}
\centering \small
\begin{tabular}{p{2.4cm}|c|c|c|c|c}
\hline
\textbf{Model}                    & \#roles & \textbf{D}$\downarrow$      & \textbf{R}$\downarrow$       & HalRate (\#hal. roles)$\downarrow$  & Granul. \\ \hline
GiT-base-coco            & 1.92             & 65.32\% (1.37)        & 14.06\% (0.29)    & 79.38\% (1.66)          & 3.93\%             \\
GiT-large-coco           & 1.94             & 65.33\% (1.36)        & 13.75\% (0.29)    & 79.09\% (1.65)          & 4.08\%             \\
GiT-base             & 0.73             & 44.05\% (0.47)& 11.98\% (0.13)& 56.03\% (0.59)& 1.8\%           \\
GiT-large                & 0.69             & \cellcolor{dustygreen}39.15\% (0.42)& 11.58\% (0.12)   & \cellcolor{dustygreen}50.63\% (0.54) & 1.89\%            \\\hline
BLIP-base-unc            & 1.44             & 61.2\% (1.01)         & 13.04\% (0.2)     & 74.23\% (1.22)          & 3.01\%             \\
BLIP-base-cond           & 2.14             & \cellcolor{dustypink}90.96\% (1.93)        & \cellcolor{dustygreen}4.22\% (0.1)& \cellcolor{dustypink}95.18 (2.03)         & 1.48\% \\
BLIP-large-unc           & 2.28             & 68.32\% (1.67)        & 13.2\% (0.31)     & 81.52\% (1.98)          & 4.38\%             \\    
BLIP-large-cond	         &2.98              & 86.6\% (2.54)         & 6.68\% (0.22)& 93.28\% (2.77)   & 2.99\% \\
BLIP2-flan-t5-xl         &1.62              & 69.26\% (1.16)        & 14.2\% (0.22)     & 83.47\% (1.38)          & 3.25\%\\
BLIP2-opt-2              &1.79              & 68.87\% (1.25)        & \cellcolor{dustypink}14.37\% (0.25)    & 83.24\% (1.51)          & 3.65\%\\\hline
ViT-GPT2                 &1.86              & 71.05\% (1.36)        & 16.46\% (0.28)    & 87.5\% (1.64)           & 3.42\%  \\\hline
Claude sonnet-L &3.9& 80.71\% (3.17)& 9.8\% (0.39)& 90.51\% (3.56)&7.1\% \\
Claude haiku-L &3.99& 80.25\% (3.29)& 10.31\% (0.38)& 90.56\% (3.67)& 6.28\%\\
 Claude sonnet-S &2.1 & 75.19\% (1.62)&  11.85\% (0.25)& 87.04\% (1.87)&5.24 \%\\
Claude haiku-S &1.85& 74.31\% (1.39)& 13.71\% (0.14) &88.02\% (1.53)& 4.99\%\\
\hline
\end{tabular}
\end{table}
In all cases, a rather high percentage of hallucinations for all semantics (objects, roles and derived phenomena) is observed; almost $1/3^{rd}$ of the caption objects present some form of hallucinations, while for roles several occurrences contain some hallucinatory inaccuracy, even exceeding 90\% in HalRate.

Replacements (\textbf{R}) represent the most common type of object edit $e$, indicating that captioners often generate hallucinated objects that are still related to the source (e.g., "daybed" instead of "couch"), rather than entirely unrelated hallucinated objects. Replacements of relevant objects are preferable to deletions for HalCECE, as the paths leading to another concept within WordNet are often shorter than the path to the root node (\textit{entity.n.01}), which corresponds to the Deletion (\textbf{D}) edit. However, for unrelated objects, where the distance between these two is greater than the cost of first deleting the one and then inserting the other, replacement is not preferable.
This is because concepts appearing in captions usually lie lower in the hierarchy, being specific enough to describe depicted objects. This level of conceptual granularity is imposed during the pre-training of captioners, which utilize descriptive captions, such as the ones of COCO or similar datasets comprising image-text pairs. On the contrary, role hierarchy is much shallower, justifying the higher number of \textbf{D} edits in comparison to \textbf{R} edits (Table \ref{tab:exp-role}). This finding is further reinforced by the fact that objects connected in $c$ often are not immediate neighbors in the ground truth, meaning that a completely new edge will need to be inserted.

A comparison between model families regarding object hallucinations (Tables \ref{tab:obj-hallucinate}, \ref{tab:obj-hallucination-2}) reveals  interesting insights: GiT variants consistently hallucinate less, achieving best results  across most metrics compared to other model families; note the \textcolor{darkerdustygreen}{colored} cells of respective Tables. On the other hand, Claude variants are accompanied with more hallucinations (note the \textcolor{darkerdustypink}{colored} cells). This may occur due to the fact that Claude models are not explicitly pre-trained on image captioning using COCO captions or similarly distributed image-text pairs, therefore they tend to hypothesize the existence of out-of-distribution concepts. 

This elevated hallucination tendency is more expected on longer captions (20-30 words), since the model is forced to be more verbose, possibly adding extraneous concepts to meet the length requirements; this is verified by the reported results, even though shorter captions are not devoid of object hallucinations as well. Additionally, as expected, longer descriptions demonstrate roughly twice the Granularity, indicating greater object coverage.
Furthermore, shorter generations are accompanied by higher over-specialization (\textbf{O}) rates, indicating that Claude models become excessively specific when attempting to condense visual information within a restricted word budget.
Another notable observation is that Haiku variants (either prompted for short or longer captions) tend to be more generic, as denoted by the inflated under-specialization (\textbf{U}) percentages in comparison to Sonnet variants, despite being prompted with the same instructions. Other than that, Haiku variants require more conceptual replacements (\textbf{R}) to assimilate the ground truth captions compared to Sonnet ones; it is possible that those \textbf{R} edits can be attributed to substitutions with concept \textit{hyponyms}, so that the \textbf{U} rates are also reduced.

A comparatively worse performance in terms of hallucinations is observed when conditional generation is employed over unconditional one in BLIP variants. This can be attributed to over-reliance over linguistic priors \cite{xiao-wang-2021-hallucination}, amplifying possible biases or noise. HalCECE is able to highlight such discrepancies regarding the generation strategy selected, suggesting straightforward mitigation strategies (in that case being the usage of unconditional caption generation). It also breaks down the source of hallucinations, as indicated in Tables \ref{tab:obj-hallucinate}, \ref{tab:obj-hallucination-2}: the rate of hallucinations (HalRate) is significantly higher than their unconditional counterparts, even though the \textbf{U} percentages are the lowest, meaning that specificity is not the culprit of hallucinations. On the contrary, the higher percentage of \textbf{D} and \textbf{R} edits denotes the presence of extraneous objects that have to be removed and substituted accordingly.

Regarding role hallucinations and comparison between models, similar trends emerge. Larger models exhibit more hallucinations overall, while GiT variants consistently produce fewer, performing best across most metrics. The primary differences across model families emerge in deletions rather than transformations, exemplified by BLIP-base-cond, which has the lowest \textbf{R} but the highest overall hallucination rate. This suggests that some models are more prone to omitting role-related information rather than altering it. Notably, Claude models demonstrate greater Granularity in role assignments, which may contribute to their higher hallucination rates. These findings align with object hallucination trends, reinforcing the idea that pre-training differences and generation strategies significantly impact hallucination tendencies across models.

Overall, it is evident that larger models cannot guarantee reduced hallucination rates. On the contrary, lower rates and fewer conceptual edits are observed in smaller captioners, such as the ones from GiT family. Even though this may sound surprising at a first glance, the source of hallucinations can be the data annotations rather than the capacity of the model itself. This means that when the visual input is ambiguous or the vision-language grounding is weak, larger models might rely more on their strong language priors, extensively stored during pre-training, thus producing fluent but unfaithful details. In the interest of image captioning, this can manifest as hallucinations -objects or actions that are statistically likely in language but \textit{not} actually present in the image.  Additionally, larger models may overfit to noisy or spurious correlations in the training data, further amplifying hallucinated content. For example, larger vision-language models may generate more detailed captions that sound plausible yet include elements unsupported by the visual evidence \cite{xiao-wang-2021-hallucination,datta2025evaluatinghallucinationlargevisionlanguage}. This suggests that the balance between visual grounding and language fluency can be more challenging to maintain as model size increases. In the following section we delve into possible discrepancies between linguistic capacity and hallucinations.

\subsection{Linguistic metrics may be misleading}
Apart from our proposed hallucination evaluation metrics, we report language generation metrics, and specifically \textit{ROUGE} \cite{lin-2004-rouge}, \textit{BLEU} \cite{papineni-etal-2002-bleu},  \textit{Google BLEU}\footnote{https://huggingface.co/spaces/evaluate-metric/google\_bleu}, \textit{Mauve} \cite{pillutla2021mauve} and perplexity (\textit{PPL}) \cite{ppl} to reveal agreements and disagreements. 

\textit{ROUGE} metrics measure recall and structural overlap between ground truth and generated captions. Specifically, \textit{ROUGE1} compares individual words (unigrams), \textit{ROUGE2} evaluates agreement between two-word sequences (bigrams), while \textit{ROUGEL} considers the longest common subsequence (LCS) between ground truth and generated text to decide upon their agreement. All these metrics are extracted by comparing the generated caption with each one of the 5 COCO captions at a time, and then obtaining their average score.
Finally, the \textit{ROUGELsum} variant regards LCS scores across multiple ground truth references (in our case being all 5 COCO captions per image), offering similar results to ROUGEL. 

\textit{BLEU} and \textit{Google BLEU} assess unigram precision between the ground truth and the generated caption, once again considering averaged values.

\textit{Mauve} provides a broader perspective on the text quality and naturalness, measuring the distributional differences between the ground truth and the generated text embeddings. It is less sensitive to exact wording and better reflects semantic similarity and stylistic variability.
In technical terms, we opt for GPT2 as the decoder to obtain embedding representations, following the default setup\footnote{https://huggingface.co/spaces/evaluate-metric/mauve}.

All those metrics range between [0,1] with higher values being better.

\textit{Perplexity (PPL)} quantifies how “surprised” a language model is when it sees the next word in a sequence, providing a measure of confidence in accurately predicting the next word. This higher confidence is associated with more predictable, fluent and coherent textual generations, reflected in lower PPL scores. A perfect PPL score equals to 1, while no upper bound exists.
\paragraph{\textbf{What is the issue with language generation metrics?}}
While these widely used metrics provide useful signals—primarily around fluency, style, and surface-level similarity—they can be misleading indicators of overall quality in text generation, often failing to capture semantic accuracy, contextual appropriateness, and hallucinations, as expressed via factual inconsistencies \cite{fischer-etal-2022-measuring,kryscinski-etal-2020-evaluating}. 

For example, n-gram overlaps reward surface-level similarity, totally excluding semantically equivalent expressions or even word ordering variability. For example, if an image contains the concept "cat", n-gram metrics will assign the \textit{same penalty} over captions that contain either the concepts "kitten" or "ship" in place of "cat". On the contrary, HalCECE will provide a significantly higher \textbf{R} cost to the "cat"$\rightarrow$"ship" edit in comparison to the "cat"$\rightarrow$"kitten" one.
Even semantically adaptive metrics, such as \textit{Mauve}, are not oriented towards factual inconsistencies, as reflected on disagreements between the visual and the linguistic modalities. This means that a caption can be perfectly natural and well-written, achieving high \textit{Mauve} scores, while also containing several objects or roles not existing in the corresponding image. Similarly, \textit{PPL} penalizes inarticulate generations but totally ignores semantic disagreements between modalities. Overall, apart from the n-gram overlap metrics, the rest are by design \textit{not explainable}; their reliance on linguistic distributions sacrifices senses of semantic interpretability, leading to obscure and dispersed evaluation practices in the first place. Finally, in all cases, linguistic metrics require ground truth captions in order to function, contrary to HalCECE which only requests standalone concepts.

Based on the above, the motivation behind our explainable and conceptual hallucination detection framework is further verified by the unsuitability and opaqueness of common text generation evaluation practices. Therefore, the language generation metrics are incapable of providing proper hallucination signals on their own, and in several cases -e.g. when n-grams are employed to measure agreement- they can even be \textit{misleading}.
These arguments will be analyzed with the support of language generation metric results, as presented in Table \ref{tab:nlp-metrics}.

\begin{table}[t!]
\caption{Language generation evaluation metrics on the $\textit{VG}\cap \textit{COCO}$ validation subset.}
\label{tab:nlp-metrics}
\centering \small
\begin{tabular}{p{2.4cm}|cccc}
\hline
\textbf{Models}         & \textbf{ROUGE1}$\uparrow$ & \textbf{ROUGE2}$\uparrow$ & \textbf{ROUGEL}$\uparrow$ & \textbf{ROUGELsum}$\uparrow$  \\ \hline
GiT-base-coco  & 0.152  & 0.021  & 0.145  & 0.145       \\
GiT-large-coco & 0.152  & 0.022 & 0.146  & 0.146        \\
GiT-base   & 0.139  & 0.01   & 0.134  & 0.134      \\
GiT-large      & 0.127  & 0.01   & 0.122  & 0.122      \\\hline
BLIP-base-unc  & 0.16   & 0.021  & 0.153  & 0.154      \\
BLIP-base-cond & 0.352 & 0.116 & 0.317 & 0.317 \\
BLIP-large-unc & 0.134  & 0.017  & 0.126  & 0.126  \\
BLIP-large-cond & 0.402 & 0.163 & 0.361 & 0.361\\
BLIP2-flan-t5-xl & 0.435 & 0.179 & 0.402 & 0.402\\
BLIP2-opt-2 & \cellcolor{dustygreen}0.44 & \cellcolor{dustygreen}0.187 & \cellcolor{dustygreen}0.404 & \cellcolor{dustygreen}0.404\\
\hline
ViT-GPT2 & 0.406 & 0.153 & 0.370 & 0.370\\
\hline
Claude sonnet-L & 0.133 & 0.008 & 0.117 & 0.117\\
Claude haiku-L & 0.141 & 0.011 & 0.125 & 0.125 \\
 Claude sonnet-S & \cellcolor{dustypink}0.062 & \cellcolor{dustypink}0.002 & \cellcolor{dustypink}0.058 & \cellcolor{dustypink}0.058\\
Claude haiku-S & 0.123 & 0.009 & 0.114 & 0.114 \\
\hline  & \textbf{BLEU} $\uparrow$  & \textbf{Google BLEU}$\uparrow$ & \textbf{Mauve}$\uparrow$& \textbf{PPL}$\downarrow$  \\ \hline
GiT-base-coco  & 0.0005 & 0.051       & 0.186 & 68.305   \\
GiT-large-coco  & 0.0005 & 0.051       & \cellcolor{dustygreen}0.192 & 63.629  \\
GiT-base  & 0.0001 & 0.027       & 0.131 & \cellcolor{dustypink}1541.317 \\
GiT-large  & 0.0001 & \cellcolor{dustypink}0.025     & 0.13  & 1475.033  \\\hline
BLIP-base-unc  & 0.0004 & 0.037       & 0.141 & 461.076  \\
BLIP-base-cond & 0.024 & 0.099 & 0.058 & 506.732 \\
BLIP-large-unc & 0.0003 & 0.033       & 0.132 & 67.632   \\
BLIP-large-cond & \cellcolor{dustygreen}0.056 & 0.133 & 0.064 & 127.578\\
BLIP2-flan-t5-xl & 0.046 & 0.132 & 0.067 & 211.738\\
BLIP2-opt-2 & 0.055 & \cellcolor{dustygreen}0.139 & \cellcolor{dustypink}0.009 & 130.29 \\
\hline
ViT-GPT2 & 0.051 & 0.131 & 0.068 & 69.605\\
\hline
Claude sonnet-L & 0.0001 & 0.029 & 0.174 & 71.307\\
Claude haiku-L & 0.0002 & 0.029 & 0.174 & \cellcolor{dustygreen}42.032 \\
 Claude sonnet-S & \cellcolor{dustypink}0.0 & 0.032 & 0.174 & 358.33 \\
Claude haiku-S & 0.0004 & 0.047 & 0.174 & 170.585 \\
\hline
\end{tabular}
\end{table}

\paragraph{\textbf{Analysis}} The results presented in Table \ref{tab:nlp-metrics} reveal interesting patterns, notably indicating that linguistic metrics are unsatisfactory overall, primarily because exact agreements with ground truth captions are not achieved in most cases. Specifically, even though n-gram-based metrics (i.e. \textit{ROUGE} and \textit{BLEU} variants) can explain their reported low scores, they lead to over-penalization of generations, since they do not respect semantical equivalence between concepts, contrary to HalCECE. On the other hand, \textit{Mauve} and \textit{PPL} are unable to explain themselves, despite being more semantically consistent, a gap that HalCECE is able to fill be breaking down the source of semantic disagreements.

Interestingly, linguistic metrics across models present some unexplainable variability. For example, BLIP2-opt-2 is one of the top-scorers regarding n-gram metrics, though it significantly fails according to \textit{Mauve}. This is somehow contradictory, since the same model presents a higher exact match capability over the rest, but also the lowest semantic agreement at the same time. This confusion is resolved via HalCECE, which places the hallucination performance of BLIP2-opt-2 somewhere in the middle in comparison to the other captioners, as demonstrated in Tables \ref{tab:obj-hallucinate}, \ref{tab:obj-hallucination-2}.

Comparisons between model families indicate that BLIP variants score higher in n-gram-related metrics (i.e. \textit{ROUGE} and \textit{BLEU} variants), revealing a comparatively increased adherence to ground truth captions. On the contrary, Claude models present the lowest scores regarding most n-gram metrics, revealing their reduced tendency to follow ground truth distributions. This fact was also reported in HalCECE results and related analysis of Tables \ref{tab:obj-hallucinate}, \ref{tab:obj-hallucination-2}, attributing the source of disagreeing semantics to their generic pre-training. Nevertheless, the percentages occurring from HalCECE are less strict, thanks to its semantic-driven foundations: hallucination rate (assimilating a recall-related scenario, where  the ratio of generated concepts over all relevant concepts is measured) reaches a maximum of 64.31\% (Claude haiku-L at Table \ref{tab:obj-hallucinate}), while \textit{ROUGE} variants, expressing recall-related agreement as well, reach up to 12.5\% of conceptual agreement according to \textit{ROUGEL}/\textit{ROUGELsum} scores of Table \ref{tab:nlp-metrics} for the same model, which equals to a minimum of 87.5\% hallucination rate. At the same time, \textit{ROUGE} scores, despite being able to highlight which concepts are responsible for the reported disagreements, they cannot suggest \textit{what needs to be changed}, in order to reach a dehallucinated state; conversely, a lookup in HalCECE recommendation prescribes that the 5.4\% of Haiku caption concepts are too generic, the 2.69\% are erroneously specific, while the 17.3\% and 44.33\% of caption concepts should be deleted and replaced respectively (Tables \ref{tab:obj-hallucinate}, \ref{tab:obj-hallucination-2}).

Finally, \textit{PPL} is highly uninformative when it comes to hallucinations: the high \textit{PPL} scores corresponding to GiT-base  captions denote significantly uncertain generations, even though the same captioner is associated with a low HalRate. On the contrary, Claude Haiku L presents the lowest \textit{PPL}, despite being one of the models associated with the highest HalRate. This inverse trend indicates that \textit{PPL} is a completely unsuitable evaluation measure with regard to hallucination detection, rendering any hallucination-related insights driven by \textit{PPL} severely misleading.

To sum up, we calculate the correlation between the linguistic metrics and object/role hallucination metrics as calculated from HalCECE. Related results are presented in Table \ref{tab:correlations-objects} for object hallucination metrics, and 
Table \ref{tab:correlations-roles} for role hallucination metrics,
denoting weak correlations (close to 0) between the two metric categories in both cases. Ultimately, we conclude that linguistic metrics cannot provide any useful information regarding the presence of hallucinations in image captioning, as detected from HalCECE.

\begin{table}[h!]
\caption{Correlation between the linguistic metrics and the object hallucination metrics provided by HalCECE. }
\label{tab:correlations-objects}
\centering \small
\begin{tabular}{l|>{\centering\arraybackslash}p{1.1cm}>{\centering\arraybackslash}p{1.2cm}cc>{\centering\arraybackslash}p{1cm}c>{\centering\arraybackslash}p{1cm}>{\centering\arraybackslash}p{1cm}>{\centering\arraybackslash}p{1.1cm}}
\hline
 &  \#obj. & \#ancest. & HalRate & Granul. & \textbf{U} & \textbf{D} & \textbf{O} & \textbf{R} & Sim. \textbf{R} \\ \hline
ROUGE1 & -0.15 & -0.06 & -0.04 & -0.05 & -0.05 & -0.03 & -0.03 & -0.03 & -0.03 \\ 
ROUGE2 & 0.05 & -0.04 & -0.15 & -0.15 & -0.09 & -0.06 & -0.02 & -0.06 & -0.02 \\ 
ROUGEL & -0.04 & -0.03 & 0.0 & 0.0 & -0.01 & -0.02 & -0.03 & -0.02 & -0.03 \\ 
ROUGELsum & -0.04 & -0.03 & -0.01 & -0.02 & -0.02 & -0.02 & -0.03 & -0.03 & -0.03 \\ 
BLEU & -0.01 & -0.03 & -0.08 & -0.08 & -0.06 & -0.05 & -0.03 & -0.05 & -0.03 \\ 
Google BLEU & -0.02 & -0.03 & -0.04 & -0.05 & -0.04 & -0.03 & -0.02 & -0.03 & -0.02 \\ 
Mauve & -0.02 & -0.02 & -0.02 & -0.02 & -0.02 & -0.02 & -0.02 & -0.02 & -0.02 \\ 
PPL & -0.06 & -0.02 & 0.03 & 0.01 & -0.01 & -0.03 & -0.05 & -0.03 & -0.05 \\ 
\hline
\end{tabular}
\end{table}

\begin{table}[h!]
\caption{Correlation between the linguistic metrics and the role hallucination metrics provided by HalCECE.}
\label{tab:correlations-roles}
\centering \small
\begin{tabular}{l|>{\centering\arraybackslash}p{1.1cm}>{\centering\arraybackslash}p{1.1cm}cc>{\centering\arraybackslash}p{1cm}c>{\centering\arraybackslash}p{1cm}>{\centering\arraybackslash}p{1cm}>{\centering\arraybackslash}p{1.1cm}}
\hline
 &  \#roles & D & R & HalRate (\#hal. roles) & Granul. \\ \hline
ROUGE1 & -0.1 & 0.05 & 0.03 & 0.04 & 0.03 \\ 
ROUGE2 & 0.18 & 0.01 & 0.07 & -0.03 & 0.02 \\ 
ROUGEL & -0.01 & 0.03 & 0.01 & 0.05 & 0.03 \\ 
ROUGELsum & 0.01 & 0.03 & 0.02 & 0.04 & 0.03 \\ 
BLEU & 0.07 & 0.02 & 0.05 & -0.01 & 0.02 \\ 
Google BLEU & 0.04 & 0.02 & 0.03 & 0.01 & 0.02 \\ 
Mauve & 0.02 & 0.02 & 0.02 & 0.02 & 0.02 \\ 
PPL & -0.09 & -0.01 & -0.05 & 0.01 & -0.03 \\ 
\hline
\end{tabular}
\end{table}

\section{Conclusion}
In conclusion, our novel HalCECE framework designed for detecting hallucinations in image captioning represents a pioneering stride towards explainable evaluation of ever evolving VL models. By delving into the hallucination mechanisms, we decompose related phenomena based on conceptual properties enabled by the incorporation of external hierarchical knowledge. Our proposed method imposes semantically minimal and meaningful edits to transit from hallucinated concepts present in captions to non-hallucinated ground truth ones, employing the explanatory power of conceptual counterfactuals. Moreover, previously overlooked role hallucinations are analyzed, revealing that widely-used image captioners tend to generate erroneous object interconnections more often than not. Overall, we view our current analysis as a crucial first step in the direction of accurately detecting hallucinations in VL models in a conceptual and explainable manner, paving the way for future hallucination mitigation strategies. 

\begin{credits}
\subsubsection{\ackname} We acknowledge the use of Amazon Web Services (AWS) for providing the cloud computing infrastructure that allowed the usage of Claude models. This work was supported by the Hellenic Foundation for Research and Innovation (HFRI) under the 3rd Call for HFRI PhD Fellowships (Fellowship Number 5537).
and also under the 5th Call for HFRI PhD Fellowships
(Fellowship Number 19268). 

\subsubsection{\discintname}
The authors have no competing interests to declare that are
relevant to the content of this article. 
\end{credits}
%
%
%

\bibliographystyle{splncs04}
\bibliography{main}
\appendix
\section{Proprietary model prompting}
\label{sec:prompts}
Regarding long generations, the prompt provided to Claude models is:
\begin{verbatim}
Provide me a descriptive caption of this image in English. 
The caption should contain between 20 and 30 words.
\end{verbatim}
As for short generations, the prompt is:
\begin{verbatim}
Provide me a short caption of this image in English in up 
to 10 words.
\end{verbatim} 
\section{Parameter count}
\label{sec:parameters}
Since no official parameter count exists for many of our captioners, we provide estimates based on public sources and model documentation, unless official information is available.  
These numbers may vary depending on the exact model version and any further fine-tuning or architectural modifications.
\begin{itemize}
\item GiT base: $\sim$110 million parameters
\item GiT large: $\sim$330 million parameters
\item BLIP base: $\sim$123 million parameters
\item BLIP large: $\sim$430–440 million parameters
\item BLIP2–opt–2.7: 2.7 billion parameters (official)
\item BLIP2–flan–t5 (XL version): $\sim$3 billion parameters
\item ViT–GPT2: $\sim$200 million parameters (combining a ViT–Base encoder and GPT-2 Small decoder)
\item Claude haiku \& Claude sonnet: Both are based on Anthropic’s Claude model, which is estimated to have $\sim$52 billion parameters
\end{itemize}

\end{document}